# Doppler velocity-based algorithm for Clustering and Velocity Estimation of moving objects

Mian Guo
College of Information Science & Electronic Engineering
Zhejiang University
Hangzhou, China
e-mail: 22060054@zju.edu.cn

Kai Zhong
College of Information Science & Electronic Engineering
Zhejiang University
Hangzhou, China
e-mail: zhong_kai@zju.edu.cn

*Xiaozhi Wang
College of Information Science & Electronic Engineering
Zhejiang University
Hangzhou, China
*corresponding author: xw224@zju.edu.cn

*Abstract*—We propose a Doppler velocity-based cluster and velocity estimation algorithm based on the characteristics of FMCW LiDAR which achieves highly accurate, single-scan, and real-time motion state detection and velocity estimation. We prove the continuity of the Doppler velocity on the same object. Based on this principle, we achieve the distinction between moving objects and stationary background via region growing clustering algorithm. The obtained stationary background will be used to estimate the velocity of the FMCW LiDAR by the least-squares method. Then we estimate the velocity of the moving objects using the estimated LiDAR velocity and the Doppler velocity of moving objects obtained by clustering. To ensure real-time processing, we set the appropriate least-squares parameters. Meanwhile, to verify the effectiveness of the algorithm, we create the FMCW LiDAR model on the autonomous driving simulation platform CARLA for spawning data. The results show that our algorithm can process at least a 4.5million points and estimate the velocity of 150 moving objects per second under the arithmetic power of the Ryzen 3600x CPU, with a motion state detection accuracy of over 99% and estimated velocity accuracy of 0.1 m/s.

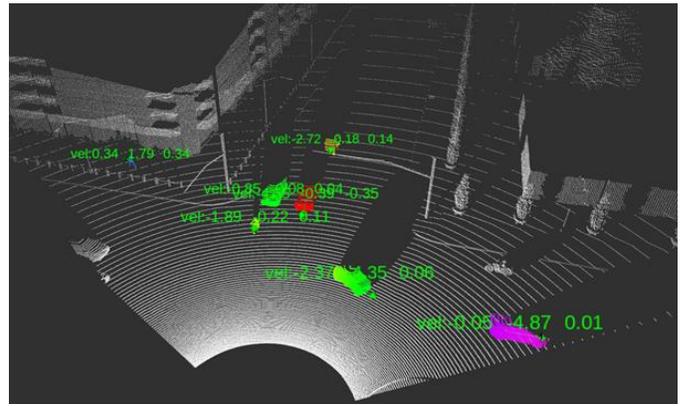

Fig. 1. The output of our method for clustering and velocity estimation. (white represents stationary and color represents movement)

## I. INTRODUCTION

Accurate localization and path planning are prerequisites for autonomous driving safety. When performing simultaneous localization and mapping (SLAM) and path planning in dynamic scenes, moving objects can have a serious impact on algorithm performance. In SLAM, we need to eliminate the moving point clouds to prevent accuracy degradation of localization and trailing shadow of mapping. As well, for path planning, we need to extract the moving point clouds in time to estimate their trajectories at the next moment and make the corresponding route design. To solve this problem, many scholars have made efforts. Based on time of flight (ToF) LiDAR, Wenjie Luo et al. proposed an end-to-end convolutional net that achieves highly accurate detection, tracking, and motion forecasting [1]. However, this method can only achieve real-time algorithm runtime, ignoring the time consumption caused by scanning. When a new object appears in the scene, ToF LiDAR often requires multiple frames (usually 3~5 frames) to estimate its motion and future trajectory, which can be deadly for high-speed vehicles. Compared to ToF LiDAR, frequency-modulated continuous-wave (FMCW) LiDAR can not only measure the position of obstacles but also obtain the Doppler velocity at each point, which provides richer sensory data. With Doppler velocity, we can easily detect the motion of objects from a single scan.

In view of the above discussion, we will take full advantage of the point clouds obtained by FMCW LiDAR which contains Doppler velocity, and propose a single-scan Doppler velocity-based algorithm for clustering and velocity estimation. By demonstrating the continuity of Doppler velocity at two adjacent points of the same object, we can easily distinguish moving objects from stationary background using the region-growing algorithm. Then we derive a formula for estimating the velocity of the FMCW LiDAR and the moving objects based on the least-squares method. Since the amount of data affects the algorithm's time consumption, we set the parameters in a reasonable way to guarantee both efficiency and accuracy. Additionally, we create an FMCW LiDAR model to verify our algorithm in CARLA which is an open-source simulator for autonomous driving research. The model parameters are set with reference to data published by several companies and the principles of the FMCW LiDAR. The result of our algorithm is shown in Figure 1. The main contributions of this paper are:

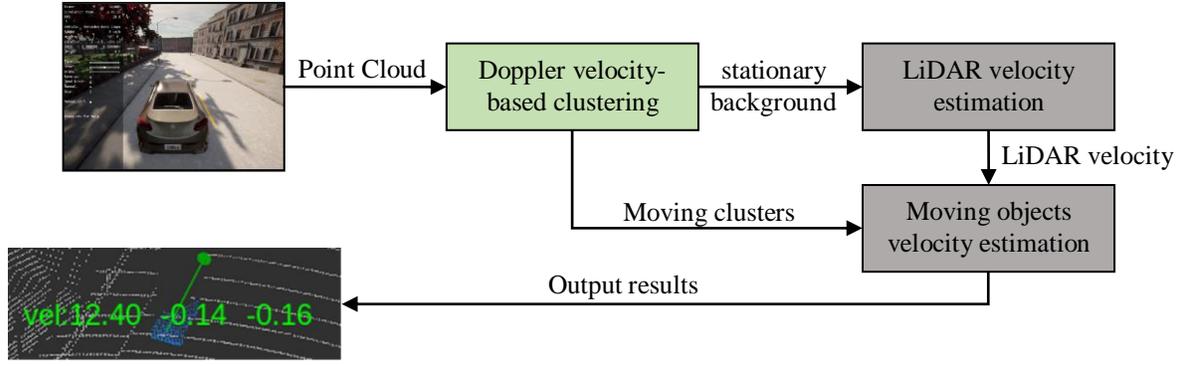

Fig. 2. The overview of the whole algorithm

- A clustering algorithm based on the continuity of Doppler velocity on the same object, that achieves real-time and stable segmentation of stationary background and moving objects.
- A velocity estimation algorithm based on Doppler velocity of point clouds that enables accurate performance through a single scan.
- An FMCW LiDAR sensor in CARLA that can be used to spawn point clouds with Doppler velocity in a simulation environment.

## II. RELATED WORK

In the field of autonomous driving, Doppler velocity is mainly measured by millimeter-wave radar [2-3]. There are many kinds of research on algorithm design using Doppler velocity. X Zhang performs dynamic gesture recognition based on micro-Doppler signatures obtained by FMCW radar [4]. Ramin Nabati incorporates Doppler velocity into the camera for accurate 3D target detection [5]. However, millimeter-wave radar is limited by the frequency band and cannot accurately recognize surrounding objects [6]. For this reason, existing methods also make very limited use of Doppler velocity. In order to achieve higher resolution, scholars tried to combine FMCW technology with LiDAR [7-11], seeking to design a sensor that can both obtain Doppler velocity and detailed image with features of obstacles. Many companies have launched FMCW LiDAR related products in recent years. For example, Blackmore used its own FMCW LiDAR for road testing and accurately distinguished the instantaneous movement of pedestrians and vehicles, which demonstrated the advantages of FMCW LiDAR for instantaneous Doppler velocity measurement. With Doppler velocity data, self-driving cars can sense the environment more reliably with lower latency, enabling safer autonomous patrols.

Each pixel point scanned by FMCW LiDAR contains a Doppler velocity, which gives us more flexibility in dealing with complex road conditions. However, up to now, the work of FMCW LiDAR has mainly concentrated on the hardware part. There are few studies on developing sensing algorithm using data from FMCW LiDAR.

## III. DOPPLER VELOCITY-BASED ALGORITHM FOR CLUSTERING AND VELOCITY ESTIMATION

### A. Algorithm Overview

We first define frames and notation that we use in this paper. We denote **L** for LiDAR frame, consistent with the vehicle coordinate system defined by the ISO. Moving objects velocity variable is $V=[x, y, z]$, where the three variables represent the velocity components on the corresponding coordinate axis in the LiDAR frame. We also assume the point cloud acquired by FMCW LiDAR at a certain scan as P, and each point $p_i \in P$ can be written as:

$$p_i = [x_i, y_i, z_i, v_i] \quad (1)$$

where the first three variables are spatial positions and the last variable is Doppler velocity.

The overview of the whole algorithm is shown in Figure 2 which consists of two parts. The algorithm receives data from FMCW LiDAR in CARLA. We segment moving objects and stationary background from the current point clouds via a region-growing algorithm. Then we estimate the velocity of the FMCW LiDAR using stationary point clouds and obtain the velocity of moving objects via moving clusters and the estimated LiDAR velocity. Velocity estimation in our algorithm is equivalent to solving a least-squares problem [12]. In addition, the velocities estimated by the algorithm are all in the current LiDAR frame.

We will introduce the following three parts: (i) Spawn point clouds in CARLA, (ii) Clustering based on Doppler velocity, (iii) Velocity estimation.

### B. Spawn point clouds in CARLA

We first introduce the principle of FMCW LiDAR about simultaneous distance and radial velocity measurement [13]. The FMCW LiDAR transmits laser waves with a triangular frequency variation over time, and the receiver receives the reflected optical signal from the measured object. The distance and speed information are obtained by measuring the beat frequency of the received signal against the original signal and bringing it into Eq. 2 and Eq. 3. The frequency relationship between the FMCW LiDAR reference and received signals is shown in Figure 3. $f_{bu}$ is the beat frequency of the rising band, $f_{bd}$ is the beat frequency of the descent band, and $f_0$ is the initial frequency of the original signal. We denote the frequency modulation period as T, the bandwidth as B, the speed of light in a vacuum as $c$, the central frequency of the original signal as $f$ and the corresponding wavelength as $\lambda$. The distance $D$ of the FMCW LiDAR from the target point can be computed as follows:

$$D = \frac{cT}{8B}(f_{bd} + f_{bu}) \quad (2)$$

the resolution of distance $\Delta D$ is computed as:

$$\Delta D = \frac{c}{2B} \quad (3)$$

the Doppler velocity of the target point can be computed using:

$$v = \frac{c}{4f}(f_{bd} - f_{bu}) = \frac{\lambda}{4}(f_{bd} - f_{bu}) \quad (4)$$

and the resolution of the Doppler velocity is computed as:

$$\Delta v = \frac{c}{fT} = \frac{\lambda}{T} \quad (5)$$

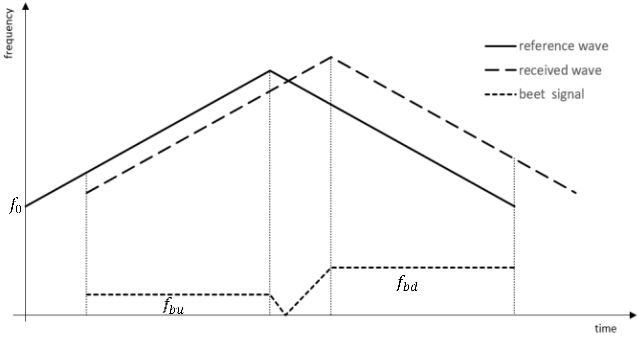

Fig. 3. Beat frequency of FMCW LiDAR

For example, we assume that the main frequency wavelength of the FMCW LiDAR is 1550 $nm$, the modulated signal has a frequency of 8~14 GHz and a period of 50 $\mu s$. Bringing the above parameters into (3) and (5), we can get the FMCW LiDAR distance resolution of 0.025m and velocity resolution of 0.031m/s. Based on the period of the modulated signal above, we can conclude that a laser can scan 20,000 points per second at most. One FMCW LIDAR contains multiple lasers for parallel scanning, which means that the imaging resolution of the FMCW LiDAR can be greatly improved. Furthermore, when multiple LIDARs are installed in a car, we can obtain image-level point cloud data. The three parameters of resolution obtained above will be used as an important reference for the FMCW LiDAR model in CARLA.

CARLA is an open-source simulator for autonomous driving research which provides many open digital assets such as sensors, urban layouts, buildings and vehicles, and supports many environmental conditions. We denote the world frame as **W** in CARLA, the velocity of models as $V_{point}^{W}$ in the world frame and $V_{point}^{L}$ in the LiDAR frame, and the velocity of LiDAR as $V_{lidar}^{W}$ in the world frame and $V_{lidar}^{L}$ in the LiDAR frame. The Doppler velocity $v_i$ obtained by FMCW LiDAR can be computed using the following equations:

$$v_i = \mathbf{e}_i \cdot (V_{point}^{L} - V_{lidar}^{L})^{\mathrm{T}} \quad (6)$$

$$V_{point}^{L} = \mathbf{R} \cdot V_{point}^{W} \quad (7)$$

$$V_{lidar}^{L} = \mathbf{R} \cdot V_{lidar}^{W} \quad (8)$$

where $\mathbf{R} \in SO(3)$ is the rotation matrix from **W** to **L**, $\mathbf{e}_i$ is the unit vector in the direction of the ray. Three of these variables $\mathbf{R}$, $V_{point}^{W}$ and $V_{lidar}^{W}$ are available via the API in CARLA, and $\mathbf{e}_i$ can be computed as the following formula:

$$\mathbf{e}_i = \frac{(x_i, y_i, z_i)}{\|x_i, y_i, z_i\|} \quad (9)$$

Using the above formulas, we then convert 3D LiDAR to the FMCW LiDAR sensor in CARLA. The main parameters of the FMCW LiDAR model we created are shown in Table I. Measurement error value setting with the same resolution.

TABLE I. PARAMETERS OF FMCW LIDAR

| Parameter | Value |
| --- | --- |
| Distance error | 0.025m |
| Velocity error | 0.031m/s |
| Angular resolution | 0.2°×0.2° |
| Field of view | 40°×30° |
| Detection range | 150m |
| Operating frequency | 10Hz |

In CARLA, we set up three LIDARs on the vehicle with a field of view of 120°×30°, and collect road data in different scenes under stationary and motion conditions.

### C. Clustering based on Doppler velocity

We assume that each point of the same object have the same velocity. In order to facilitate the calculation, we remove the superscript and change Eq. 6 as:

$$v_i = \|\mathbf{v}_{model} - \mathbf{v}_{lidar}\| \cdot \cos\theta_i \quad (10)$$

where $\theta_i$ is the angle between the relative velocity direction of the point and the direction of the ray. Then we can calculate the Doppler velocity difference $\Delta v_{same}$ between two adjacent points $a$ and $b$ on the same objects as:

$$\begin{aligned}\Delta v_{same} &= |v_a - v_b| \\ &= \|\mathbf{v}_{model} - \mathbf{v}_{lidar}\| \cdot |\cos\theta_a - \cos\theta_b|\end{aligned} \quad (11)$$

also due to

$$\begin{aligned}|\theta_a - \theta_b| &\approx \theta_{res} \\ &\approx \pi \times \frac{0.2°}{180°} \\ &\approx 0.0034 \; rad\end{aligned} \quad (12)$$

we can perform a Taylor expansion for $\cos(\cdot)$ at $\theta_b$ to obtain the following equation:

$$\Delta v_{same} = \|\mathbf{v}_{model} - \mathbf{v}_{lidar}\| \cdot \left| -\frac{\theta_{res}^2}{2}\cos\theta_b - \theta_{res}\sin\theta_b + \frac{\theta_{res}^3}{3!}\sin\theta_b + o(\theta^3) \right| \quad (13)$$

bringing Eq. 12 into Eq. 13, we can obtain the equation as:

$$\begin{aligned} \Delta v_{same} &\approx \|\mathbf{v}_{model} - \mathbf{v}_{lidar}\| \cdot \left| -\frac{\theta_{res}^2}{2}\cos\theta_b - \theta_{res}\sin\theta_b \right| \\ &\approx \|\mathbf{v}_{model} - \mathbf{v}_{lidar}\| \cdot \left| \theta_{res} \cdot \cos(\theta_b - \arctan\frac{2}{\theta_{res}}) \right| \quad (14) \\ &\leq (\|\mathbf{v}_{model}\| + \|\mathbf{v}_{lidar}\|) \cdot \theta_{res} \end{aligned}$$

The Doppler velocity difference $\Delta v_{diff}$ between two adjacent points of different objects (one of which is stationary) can be obtained by the same way. The $\Delta v_{diff}$ can be expressed by the following formula:

$$\Delta v_{diff} \approx \|\mathbf{v}_{model}\| \cdot |\cos\varphi_{model}| \quad (15)$$

where $\varphi_{model}$ is the angle between the absolute velocity direction of the point and the direction of the ray. Usually, we can assume that the maximum speed of the moving objects in an urban area is 90 km/h, i.e. 25 m/s. Then we can obtain $\Delta v_{same} \leq 0.17$ using Eq. 14. In order to separate moving objects from the stationary background, we need to satisfy $\Delta v_{diff} \geq 0.17 \geq \Delta v_{same}$. From Eq. 15, it can be concluded that if we want to identify objects with speeds above $v$, a certain area has to be lost where we cannot distinguish the motion of the point cloud. For example, if we want to identify an object with a speed of 5 m/s, the blind area can be calculated roughly by Eq. 15 as $\{\varphi_{model} \in [-92°, -88°] \cup [88°, 92°]\}$.

According to the properties of $\Delta v_{same}$ and $\Delta v_{diff}$, the motion and stationary points within a certain condition can be easily separated using region growing algorithm by setting a threshold value. Meanwhile, we can draw the following conclusions: with the same motion direction, the faster the object is, the easier it is to be separated out; with the same speed, the farther the angle between the motion direction and the LiDAR ray direction is from ±90°, the easier the object is to be separated out. Due to the fact that the number of points of moving objects in a single scan is usually much less than the number of stationary points, we denote the largest number of point cloud as stationary point cloud $P_s$, and the rest as moving point cloud $P_m$. With this method, the point cloud $P$ at a certain frame can be clustered as follows:

$$\begin{aligned} P &= \{P_s, P_m\} \\ where: P_m &= \{P_{m1} \cup P_{m2} \cup \ldots \cup P_{mk}\} \end{aligned}$$

$P_{mk}$ is the point cloud of the *k*th moving object. The Doppler velocity-based clustering algorithm is shown in Algorithm 1.

---

**Algorithm 1:** clustering based on Doppler velocity

**Input:** point cloud P
**Output:** $P_s$, $P_m$
1  Store P into the matrix *Mat*;
2  Set the initial point $p$, set threshold $v_{th}$;
3  Initialize $P_s$, $P_m$;
4  k = 0, $P_{mk} = \{Mat_{(0,0)}\}$;
5  **while** (1)
6     **while** unprocessed point $Mat_{(x,y)}$ exists in $P_{mk}$
7        **if** $abs(Mat_{(x\pm1,y\pm1)}.vel - Mat_{(x,y)}.vel) < v_{th}$ **then**
8           $P_{mk} += Mat_{(x\pm1,y\pm1)}$;
9           $Mat_{(x\pm1,y\pm1)}.processed = 1$;
10       **end**
11    **end**
12    **if** unprocessed point $Mat_{(x,y)}$ exists in P **then**
13       k++;
14       $P_{mk} = \{Mat_{(x,y)}\}$;
15    **else**
16       *break*;
17    **end**
18 **end**
19 $P_s = Max\{P_{m1}, \ldots, P_{mk}\}$, $P_m = \{P_{m1}, \ldots, P_{mk}\} - P_s$;

---

*D. Velocity estimation*

We obtain the stationary background data $P_s$ and the set of motion point clouds $P_m$ by Algorithm 1. We denote the unit direction vector for each point $\mathbf{p}_i = (x_i, y_i, z_i, v_i) \in P_s$ and FMCW LiDAR velocity in LiDAR frame as:

$$\mathbf{e}_i = \frac{(x_i, y_i, z_i)}{\|x_i, y_i, z_i\|} = (x_{ei}, y_{ei}, z_{ei}),\ i=1,2,\ldots,n \quad (16)$$

$$\mathbf{V}_{self} = (v_x, v_y, v_z) \quad (17)$$

We define $(\mathbf{A}, \mathbf{V})$ as a pair of observations where:

$$\mathbf{V} = [v_1, v_2, \ldots, v_n]^T \quad (18)$$

$$\mathbf{A} = [\mathbf{e}_1, \mathbf{e}_2, \ldots, \mathbf{e}_n]^T = \begin{bmatrix} x_{e1} & y_{e1} & z_{e1} \\ x_{e2} & y_{e2} & z_{e2} \\ \vdots & \vdots & \vdots \\ x_{en} & y_{en} & z_{en} \end{bmatrix} \quad (19)$$

From Eq. 6, $(\mathbf{A}, \mathbf{V})$ satisfies the following function:

$$\mathbf{V} = f(\mathbf{A}, \mathbf{V}_{self}) = \mathbf{A} \cdot (0 - \mathbf{V}_{self})^T \quad (20)$$

the optimal estimation of the parameter $\mathbf{V}_{self}$ of the function $f(\mathbf{A}, \mathbf{V}_{self})$ is equivalent to calculating the parameter $\mathbf{V}_{self}$ when the objective function:

**Algorithm 2:** velocity estimation

**Input:** $P_s$, $P_m$
**Output:** $vectorArray_{vel}$
/* P.downSample($num$) : randomly take $num$ points. */
/* velEstimate(P) : estimation by Ceres Solver using Eq. 21 and Eq. 22 */

1. Set the max number $num_{th}^s$;
2. $V_{self}$ = vel_estimate($P_s$.downSample($num_{th}^s$));
3. $vectorArray_{vel} = \{\}$, set the max number $num_{th}^m$;
4. **for** $P_{mk}$ in $P_m$ **do**
5.    **if** $num_{th}^m < P_{mk}.num$ **then**
6.       $P_{mk}$.downSample($num_{th}^m$);
7.    **end**
8.    $V_{model}$ = vel_estimate($V_{self}$, $P_{mk}$);
9.    $vectorArray_{vel}$ += $V_{model}$;
10. **end**

$$L(\mathbf{V}, f(\mathbf{A}, \mathbf{V}_{self})) = \mathbf{V} + \mathbf{A} \cdot \mathbf{V}_{self}^{\mathbf{T}} \quad (21)$$

takes the minimum value. Thus, we transform velocity estimation into a least-squares problem, which can be solved by many methods, such as QR decomposition [14] and SVD [15], etc. In the same way, we can also transform the velocity estimation of the target moving object into a least-squares problem:

$$L(\mathbf{V}, f(\mathbf{A}, \mathbf{V}_{model})) = \mathbf{V} + \mathbf{A} \cdot \mathbf{V}_{self}^{\mathbf{T}} - \mathbf{A} \cdot \mathbf{V}_{model}^{\mathbf{T}} \quad (22)$$

The disadvantage of using this method for velocity estimation is obvious that the computational time increases significantly when the number of point clouds is too large. Thus, we need to set a maximum value to constrain the number of data for the least-squares problem. Based on the experiments in the later section, we arrive at an optimal upper limit of 200 for model velocity estimation to achieve a balance between accuracy and time consumption. We use Ceres Solver to solve the least-squares problem which is an open-source C++ library developed by Google [16]. The velocity estimation is shown in Algorithm 2.

## IV. EXPERIMENTS

TABLE II. INFORMATION OF DATASET

| Dataset | Scans | Total number of motion points | Time | Movement Status |
|---|---|---|---|---|
| intersection | 135 | 473271 | 13.5s | Stationary |
| T-intersection | 98 | 463957 | 9.8s | Stationary |
| Straight road | 91 | 122893 | 9.1s | Moving |
| Turn & Straight road | 151 | 1256177 | 15.1s | Moving |

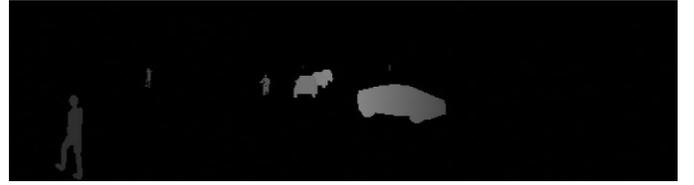
(a) Intersection

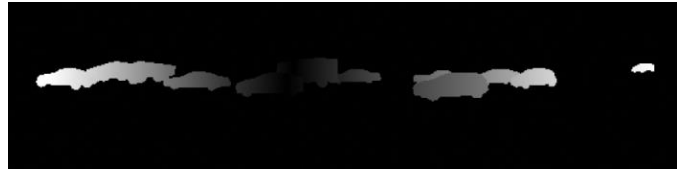
(b) T-intersection

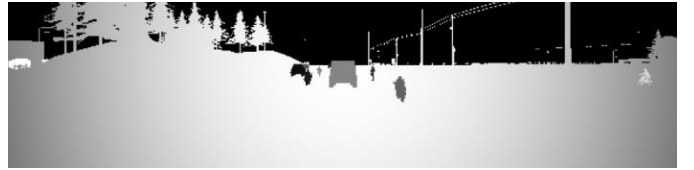
(c) Straight road

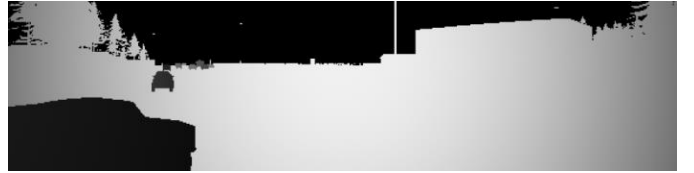
(d) Turn & Straight road

Fig. 4. Camera view of the absolute Doppler velocity

The experiments in this paper are conducted with an AMD Ryzen 3600x CPU using the robot operating system (ROS) in Ubuntu 18.04. The algorithms are implemented in C++. We record four datasets in CARLA containing two stationary scenes and two motion scenes, which are *intersection*, *T-intersection*, *straight road* and *turn & straight road*. The intersection dataset records straight ahead and turning vehicles & walking pedestrians. The T-intersection dataset records lateral moving vehicles. The straight road and turn & straight road record forward, opposite and sideways moving vehicles. The datasets information is shown in Table II.

The camera view of the absolute Doppler velocity measured by FMCW LiDAR is shown in Figure 4, which provides a more visual representation of the FMCW LiDAR scan in four scenarios. The black areas in the four scenes are points with a Doppler velocity of 0 or empty points.

### A. Clustering based on Doppler velocity

The motion points are attributed with positive value, while the stationary points are negative. We denote TP for correctly classified positive classes, TN for correctly classified negative classes, FP for incorrectly classified positive classes, and FN for incorrectly classified negative classes. We use the following formula for the evaluation metrics of the algorithm:

$$Precision = \frac{TP}{TP + FP} \quad (23)$$

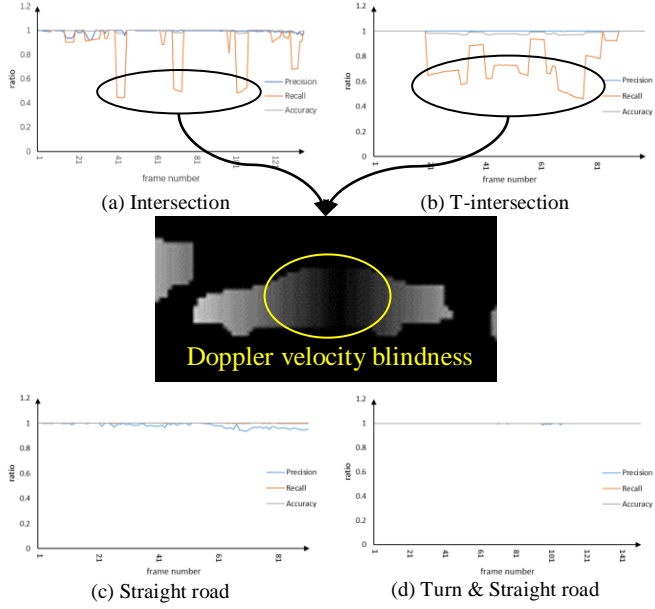

(a) Intersection  (b) T-intersection

(c) Straight road  (d) Turn & Straight road

Fig. 5. Performance of clustering

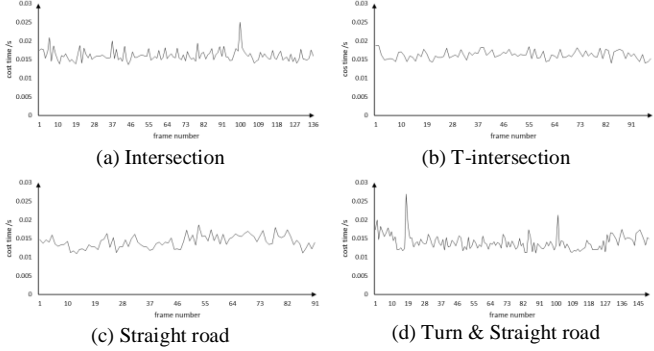

(a) Intersection  (b) T-intersection

(c) Straight road  (d) Turn & Straight road

Fig. 6. Time consumption of clustering

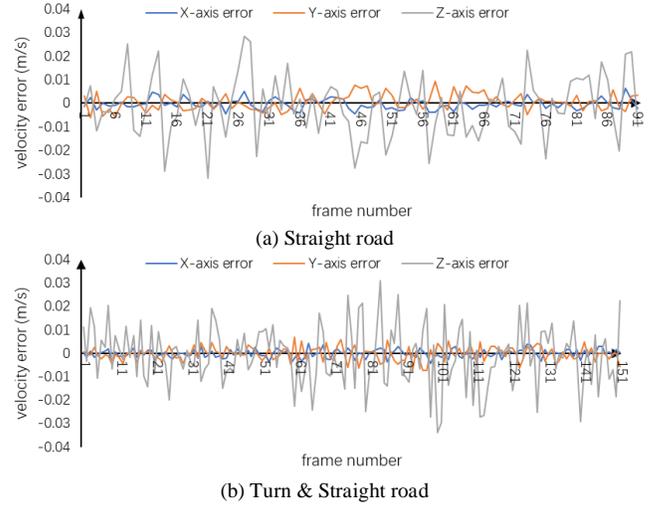

(a) Straight road

(b) Turn & Straight road

Fig. 7. LiDAR velocity estimation errors

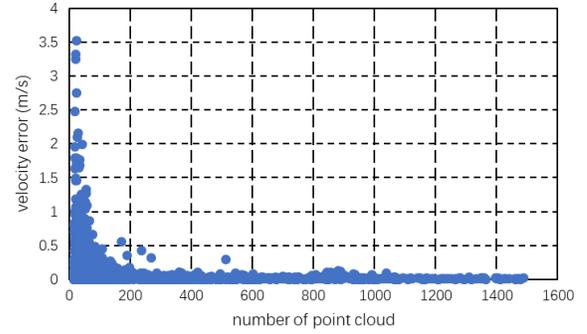

Fig. 8. Model velocity estimation errors

$$Recall = \frac{TP}{TP+FN} \quad (24)$$

$$Accuracy = \frac{TP+TN}{TP+TN+FP+FN} \quad (25)$$

Testing in different scenarios reflects the strengths and weaknesses of the Doppler velocity-based clustering. The performance of clustering is shown in Figure 5. The accuracy of the algorithm in distinguishing between moving and stationary points remains largely above 99% in the four scenarios. However, in some special cases, there is a sudden drop in the algorithm's recall rate. When the velocity direction of the point cloud is close to perpendicular to the ray direction of the FMCW LiDAR, the point cloud is moving though its characteristics in terms of Doppler velocity are similar to that of stationary points so that in the process of clustering, the algorithm will incorrectly classify the point cloud into stationary group. We refer to this situation as Doppler velocity blindness. Doppler velocity blindness rarely occurs on a straight or curved road because the objects moving on the road are generally in forward or opposite direction. Therefore in (c) and (d) of Figure 5, recall can be maintained at a high level all the time.

The time required to process each frame is shown in Figure 6. The Doppler velocity-based clustering distinguishes between stationary and moving points by performing a single traversal of the point cloud matrix. Thus, the time taken by the algorithm is only related to the size of the matrix, and the time consumption is always low regardless of the number of moving points in the scene. As can be seen from Figure 6, the running time of the algorithm is stable within 20 *ms*, which indicates that the algorithm can process at least 4.5 million points per second with the arithmetic power of the AMD Ryzen 3600x CPU.

*B. Velocity estimation*

We show in Figure 7 the estimation error of the LiDAR velocity in the motion state. When performing LiDAR velocity estimation, the velocity is more accurate on the X and Y axes, with errors remaining within 1 cm/s, while on The Z axis the estimation errors remain within 3 cm/s. Since the field of view used in this experiment is 120°×30°, there is more data on the lateral field of view compared to the longitudinal field of view and stronger constraints on the velocity estimates in the X and Y directions.

For the velocity estimation of moving objects, we collate the number of moving object point clouds and their velocity estimation errors in four scenarios and plot a scatter plot, as shown in Figure 8. We can see that the number of moving object point clouds can greatly affect the velocity estimation accuracy. When the number of point cloud exceeds 200, the velocity estimation error almost stays within 10 cm/s. For the same object, the closer the object is to the LIDAR, the higher the velocity estimation accuracy; for different objects, the larger the area scanned by the LiDAR at the same distance, the higher the velocity accuracy.

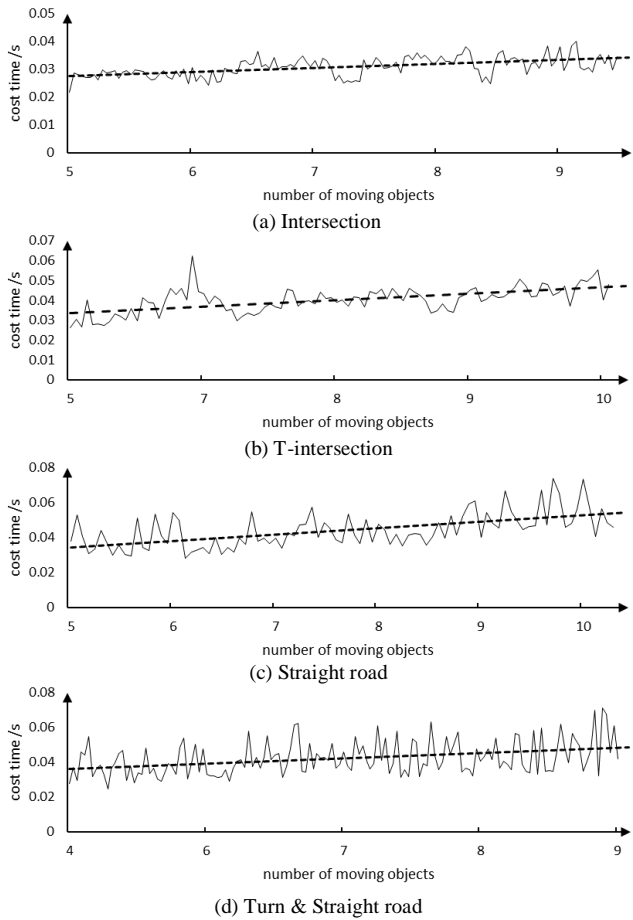

Fig. 9. Time consumption of velocity estimation

The total time of the velocity estimation algorithm is the sum of the time consumption of LiDAR and moving objects velocity estimation. The time taken by moving objects velocity estimation depends on the number of moving objects, so the time consumption is a linearly increasing function of the number of moving objects, as shown in Figure 9. The value of the trend line when the number of moving objects is 0 is the time consumption of LiDAR velocity estimation. Based on this trend line, the algorithm can estimate the velocity of up to 15 moving objects per scan in real time with the AMD Ryzen 3600x CPU.

## V. Conclusion and Discussion

We have proposed Doppler velocity-based clustering and velocity estimation for motion state recognition and trajectory prediction. This is the first use of Doppler velocity obtained by FMCW LiDAR. By comparing the Doppler velocity difference between two adjacent points in the point cloud matrix, the Doppler velocity-based clustering achieves a performance of 50 frames per second with a detection accuracy of over 99%. By solving the least squares problem, the velocity estimation achieves a 0.03 m/s LiDAR velocity estimation error and a 0.1 m/s object velocity estimation error. These two algorithms provide great convenience in the recognition, tracking and prediction of moving objects. Future work will focus on eliminating Doppler velocity blindness and improving robustness.